\documentclass[runningheads]{llncs}

\usepackage{eccv}

\usepackage{eccvabbrv}

\usepackage{graphicx}
\usepackage{booktabs}
\usepackage{amssymb} 
\usepackage{pifont} 
\usepackage{multirow} 
\usepackage{tikz}

\usepackage[accsupp]{axessibility}  

\usepackage{hyperref}

\usepackage{orcidlink}

\begin{document}

\title{Forecasting Future Videos from Novel Views\\ via Disentangled 3D Scene Representation} 

\titlerunning{Forecasting Future Videos from Novel Views}

\author{Sudhir Yarram\inst{1}\orcidlink{0000-0002-7986-3778} \and
Junsong Yuan\inst{1}\orcidlink{0000-0002-7901-8793}}
\authorrunning{Y.~Sudhir et al.}

\institute{State University of New York at Buffalo\\
\url{https://skrya.github.io/projects/ffn-dsr}}

\newcommand{\largegap}[1]{{\vspace{-20pt}}}
\newcommand{\gap}[1]{{\vspace{-9pt}}}
\newcommand{\lessgap}[1]{{\vspace{-4pt}}}
\newcommand{\lowestgap}[1]{{\vspace{-2pt}}}

\maketitle

\begin{abstract}
Video extrapolation in space and time (VEST) enables viewers to forecast a 3D scene into the future and view it from novel viewpoints. Recent methods propose to learn an entangled representation, aiming to model layered scene geometry, motion forecasting and novel view synthesis together, while assuming simplified affine motion and homography-based warping at each scene layer, leading to inaccurate video extrapolation. 
 Instead of entangled scene representation and rendering, our approach chooses to disentangle scene geometry from scene motion, via lifting the 2D scene to 3D point clouds, which enables high quality rendering of future videos from novel views. To model future 3D scene motion, we propose a disentangled two-stage approach that initially forecasts ego-motion and subsequently the residual motion of dynamic objects (e.g., cars, people). This approach ensures more precise motion predictions by reducing inaccuracies from entanglement of ego-motion with dynamic object motion, where better ego-motion forecasting could significantly enhance the visual outcomes. Extensive experimental analysis on two urban scene datasets demonstrate superior performance of our proposed method in comparison to strong baselines.

\end{abstract}
\gap{}
\gap{}




\section{Introduction}
\label{sec:intro}
\vspace{-6pt}
\begin{figure*}[t!]
    \centering
    \includegraphics[width=1.0\textwidth]{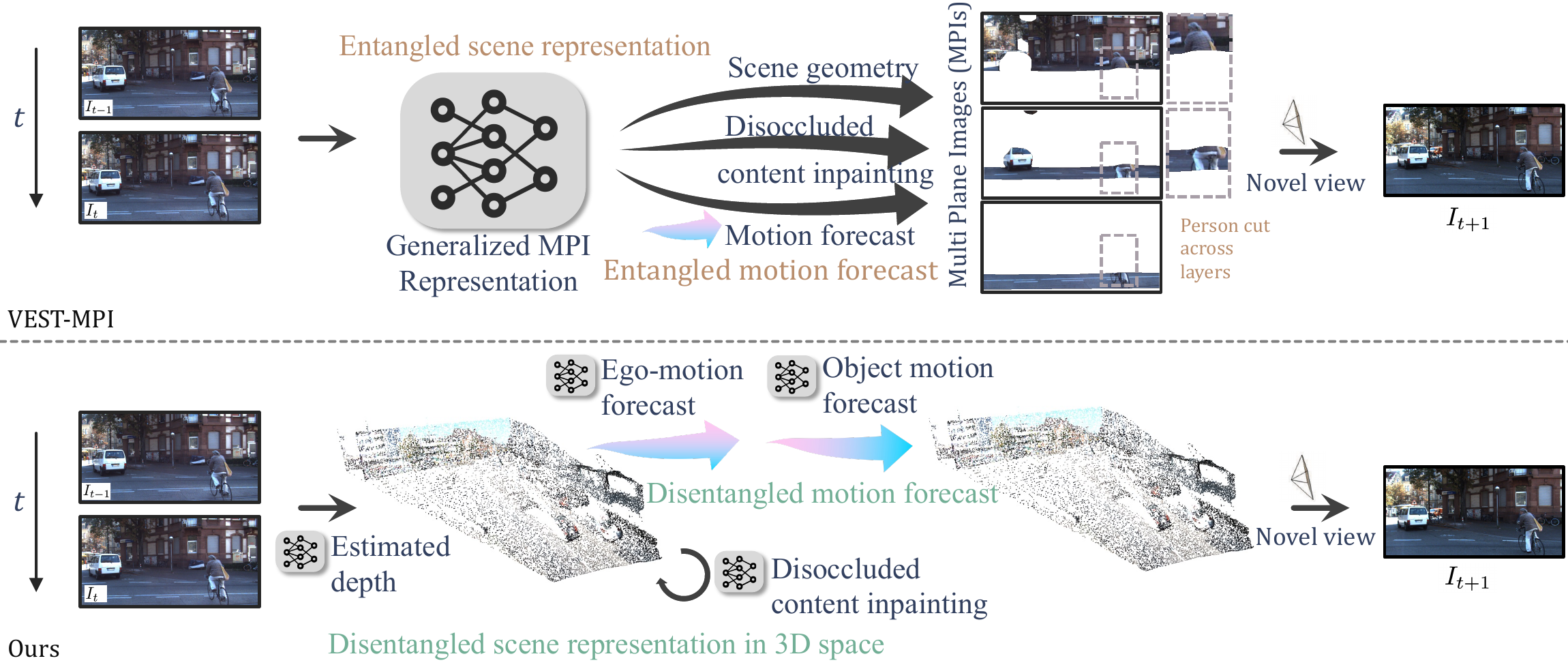}
    \vspace{-8pt}
    \caption{\textbf{Comparisons of VEST approaches.} Compared to VEST-MPI \cite{zhang2022video}, our method features: \textbf{(1) Disentangled 3D geometry and motion representation:} While VEST-MPI \cite{zhang2022video} relies on an entangled layered MPI representation with simplified affine motion and homography-based warping, we employ depth maps to transform 2D images into 3D point clouds, disentangling scene geometry from motion for high-quality rendering from novel viewpoints; \textbf{(2) Disentangled ego-motion and object motion forecast:} Departing from VEST-MPI's simultaneous modeling of ego-motion and object motion, we adopt a disentangled two-stage forecasting approach. Our approach first predicts ego-motion, then addresses residual object motion. This separation allows our model to predict 3D motion more accurately, improving the accuracy of 3D motion forecasts. }
    \vspace{-12pt}
    \label{fig:intro}
\end{figure*}
The Video Extrapolation in Space and Time (VEST) task, as introduced in \cite{zhang2022video}, combines the tasks of novel view synthesis and video prediction. VEST involves extrapolating scenes to novel views while simultaneously being able to forecast future videos. Addressing VEST requires solving intricate challenges such as estimating scene geometry, forecasting future motion, and synthesizing disoccluded content due to object movement and novel viewpoints. 

Modeling the complex relationship between view changes ({\it geometry}) and future scene ({\it motion}) for an entire scene presents a significant challenge. To address this complexity, existing approaches like VEST-MPI \cite{zhang2022video} leverage generalized multi-plane image (MPI) representation. This technique simplifies the task by breaking down images into layers based on fixed depth, facilitating the modeling of view changes and future motion on a per-layer basis.

Despite the efficacy of this MPI-based approach, it has two major limitations, as illustrated in Fig.~\ref{fig:intro}. First, the learned representation of each layer tends to be highly entangled. It attempts to simultaneously model scene geometry, future scene motion and the synthesis of disoccluded content. This entanglement arises from the assumptions that all pixels within a layer can render novel view through simple homography-based warping, and future scene motions can be approximated via affine transformations alongside flow-based warping. However, these assumptions often fall short in complex scenes, such as urban driving scenarios, where the sophisticated geometry and the diverse motions of objects within each layer challenge these simplifications. Consequently, this can result in blurred visual content, especially in long-term forecasts.  

Second, the layered MPI representation faces challenges in accurately modeling future 3D motion, primarily due to the complexity inherent in predicting both ego-motion and dynamic object motion within each layer. Ego-motion affects all observed elements in the scene and can become entangled with the irregular motion of dynamic objects, potentially causing inaccuracies in the motion prediction. The precise ego-motion forecasting is crucial since even slight errors can lead to significant deviations in the visual outcome. It thus underscores the importance of disentangling these motion components to achieve more accurate motion predictions. 

To address these challenges, we propose an approach that models 3D scene geometry within a continuous space of point clouds rather than a layered MPI. This continuous scene representation aids in the disentanglement of scene geometry from scene motion, enabling the rendering of scene from novel camera perspectives, thereby achieving high-fidelity motion forecasting with photorealistic synthesis of future videos. The key to our approach is to leverage estimated depth map and lift the scene representation from the 2D RGB images to 3D point clouds. Each 3D point is defined by its 3D location and a learned appearance feature. These point clouds can effectively account for 3D scene motion by displacing 3D location of the corresponding point based on its physical motion flow. Finally, we can use sophisticated 3D-to-2D rendering techniques \cite{ravi2020accelerating} to process these point clouds, to synthesize the future frame from a novel view, instead of simple homography-based warping. 

To accurately forecast future motion, moving away from simultaneously estimating ego-motion and object motions used in VEST-MPI \cite{zhang2022video}, we propose a disentangled two-stage 3D motion forecasting approach. Our two-stage approach begins by forecasting camera ego-motion and subsequently addresses the residual motion of dynamic objects (\eg car, people), thereby modeling future scene motion. Since static background motion in past frames can estimate future ego-motion, we introduce an ego-motion forecasting module that analyzes static background elements to forecast ego-motion. Subsequently, we introduce a multi-scale object forecasting module to address the residual motion of objects. This module analyzes the scene elements across past frames to predict residual object motion. Utilizing the forecasted 3D motion, we relocate the point clouds, creating a future 3D scene and then synthesizing the future frame.

In summary, our contributions include: (1) a novel approach that leverages explicit 3D scene geometry to address VEST by disentangling 3D scene geometry from scene motion, enabling the rendering of scene from novel views, and thereby achieving high-fidelity scene forecasting with photorealistic synthesis of future videos. (2) Disentangled two-stage approach for future motion forecast that first forecasts ego-motion and then the residual motion of dynamic objects, in order to achieve high quality video extrapolation in space and time.

Our method outperforms baselines in VEST, video prediction, and novel view synthesis on benchmark datasets such as KITTI \cite{geiger2013vision} and Cityscapes \cite{cordts2016cityscapes}.



\begin{figure*}[t]
    \centering
    \includegraphics[width=1.0\textwidth]{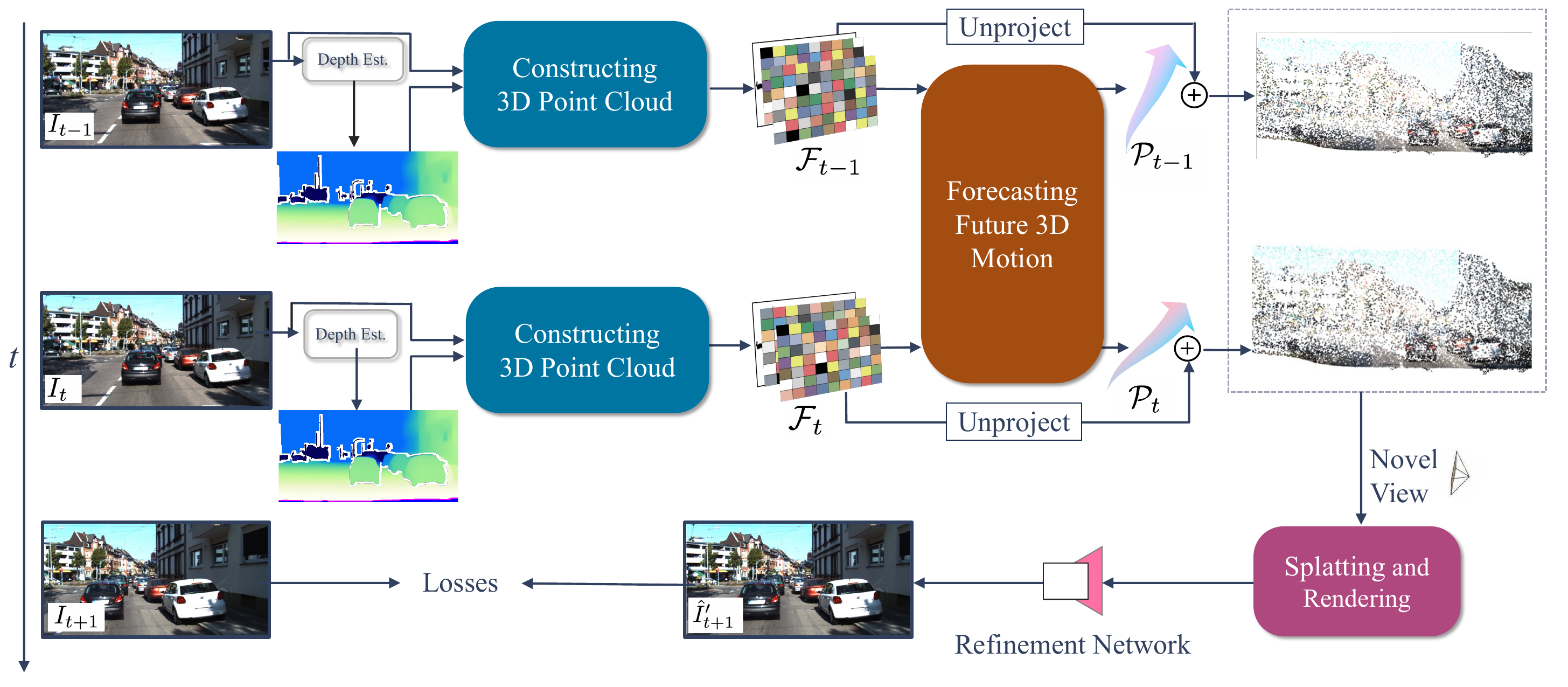}
    \vspace{-12pt}\lessgap{}
    \caption{\textbf{Method overview.} Our framework aims to forecast a 3D scene into the future and view it from novel viewpoints. It comprises three primary steps  \textbf{(1) Constructing 3D point clouds:} Starting with two past frames as the input, we construct per-frame 3D point clouds. (i) The process for each frame involves depth estimation, dis-occlusion handling via inpainting, and feature extraction to finally generate what we refer to as feature layer. (ii) The point-wise features in this feature layer are then lifted into 3D space using corresponding depth values, generating 3D point clouds. This process is performed on both $\mathbf{I}_{(t-1)}$ and  $\mathbf{I}_{(t)}$ to obtain feature layers $\mathcal{F}_{(t-1)}$ and $\mathcal{F}_{(t)}$ and point clouds $\mathcal{P}_{(t-1)}$ and $\mathcal{P}_{(t)}$.  \textbf{(2) Forecasting future 3D motion}: We leverage the feature layers $\mathcal{F}_{(t-1)}$ and $\mathcal{F}_{(t)}$ to forecast future 3D motion for each of the point clouds. This forecasted 3D motion allows us to update the positions of point clouds $\mathcal{P}_{(t-1)}$ and $\mathcal{P}_{(t)}$ to their new, forecasted locations.  \textbf{(3) Splatting and Rendering:} A point-based renderer processes these motion-adjusted point clouds through 3D-to-2D splatting to generate feature maps. Finally, refinement network takes these rendered feature maps and decodes them to synthesize a novel view $\hat{I}'_{(t+1)}$ based on the target viewpoint.}
    \vspace{-12pt}
    \label{fig:overview}
\end{figure*}
\section{Related Work}
\label{sec:related_work}
\vspace{-10pt}



\paragraph{Video prediction.}
Previous video prediction (VP) methods vary from unconditional synthesis \cite{clark2019adversarial, denton2018stochastic, finn2016unsupervised,franceschi2020stochastic, saito2020train, kumar2019videoflow, tian2021good} to conditional prediction tasks \cite{le2023waldo, bei2021learning, hu2023dynamic, wu2020future, geng2022comparing, schmeckpeper2021object, wang2018video}. These conditional prediction methods use diverse strategies to forecast motion. Some decompose videos using object-centric \cite{wu2020future}, semantic-aware \cite{bei2021learning} and motion-aware \cite{villegas2017decomposing} approaches to forecast motion. DMVFN \cite{hu2023dynamic} proposes to handle diverse motion scales of objects through a pyramidal approach \cite{ranjan2017optical, sun2018pwc}, while \cite{le2023waldo} forecasts the trajectory of object through sparse control points. However, these methods work in either the 2D or quantized 3D domain, facing challenges in modeling occlusions. In contrast, our approach employs 3D motion forecasting, overcoming these limitations by incorporating the depth dimension. This integration enables for accurate estimates of the object motion patterns as they move in and out of each other's paths. 
\gap{}
\paragraph{Novel view synthesis.} Novel view synthesis aims to reconstruct unseen viewpoints from a set of input 2D images and their corresponding camera poses. Recent neural rendering methods have achieved impressive synthesis results \cite{liu2020neural, tewari2020state, wang2021ibrnet, mildenhall2021nerf, park2019deepsdf, sitzmann2019scene, wang2021ibrnet, li2023dynibar}. However, these methods typically assume the availability of dense views as input, a condition rarely met in practical scenarios. Moreover, their primary focus lies in interpolating across frames rather than extrapolation.  Therefore, we consider methods capable of handling sparse views as input.

Among these, several works \cite{han2022single, li2021mine, li2022symmnerf, tucker2020single, tulsiani2018layer, xu2022sinnerf, yu2021pixelnerf} aim to infer the 3D structure of scenes by learning to predict a scene representation from a single image. Yet, these methods lack the ability to forecast into the future.
\gap{}
\paragraph{Depth-based 3D scene representation.} Recent ``3D photography'' methods \cite{shih20203d, li20233d, jampani2021slide, khan2023tiled} demonstrate how estimated depth can bring 2D images to life for a 3D viewing experience. For instance, 3D Photo \cite{shih20203d} employs monocular depth mapping to generate layered depth images (LDIs) \cite{peng2022mpib}, enabling contextually aware color and depth inpainting. These methods benefit by disentangling 3D scene geometry from camera pose. However, unlike our approach, these methods assume a static scene. Among these, 3D Moments \cite{wang20223d} and Li et al.'s work \cite{li20233d} are most related to our approach as they achieve space-time view synthesis. However, 3D Moments \cite{wang20223d} focuses on space-time interpolation across frames and Li et al \cite{li20233d} focuses on animating a static scene. In contrast, our approach conditions on past frames to forecast future scene motion.



\section{Our Method}
\vspace{-6pt}
\label{sec:method}

Given two consecutive frames from a video sequence, $\mathbf{I}_{(t-1)}$ and $\mathbf{I}_{(t)}$ as the input, our goal is to forecast a future frame $\mathbf{\hat{I}'}_{(t+1)}$ rendered from a novel viewpoint. Our framework is illustrated in Fig.~\ref{fig:overview}. The output of our framework is a future video featuring high-fidelity video forecast with high-quality novel view synthesis.

Initially, we generate an estimated depth map for the input frames.  Subsequently, we construct 3D point clouds by first addressing potential dis-occlusions in each frame through inpainting, followed by feature map extraction and their unprojection into 3D space, based on depth values  (Sec.~\ref{sec:3d_scene_representation}). For future 3D motion prediction, since the 3D motion would result from both camera motion and dynamic object motions, we disentangle these 3D motions via a two-stage approach that initially forecasts ego-motion by leveraging static background regions across frames, and subsequently predicting residual motion for dynamic objects (e.g., cars, persons) (Sec.~\ref{sec:forecasting}). With these forecasts, we update the point cloud's location to reflect the anticipated 3D motion. Finally, we project these modified 3D point clouds into 2D space, according to the target viewpoint, to synthesize the future frame (Sec.~\ref{sec:rendering_NVS}). 

Our framework operates in a fully end-to-end differentiable manner, incorporating ego-motion forecasting modules and object motion forecasting modules, and directly supervises the training process based on the $L1$ loss and perceptual losses \cite{simonyan2014very}, which can be obtained by comparing the predicted future frame $\mathbf{\hat{I}'}_{(t+1)}$ and the ground truth future frame $\mathbf{I}_{(t+1)}$. 
 
\vspace{-5pt}
\subsection{Disentangled 3D Scene Representation}
\label{sec:3d_scene_representation}
\vspace{-6pt}

Instead of quantizing a 3D scene into multiple depth layers as in~\cite{zhang2022video}, we introduce an approach to model the geometry of a 3D scene as a continuous space of point clouds. This is achieved via lifting scene from 2D RGB images to 3D through estimated depth map. This simple yet effective treatment provides feasibility to disentangle scene geometry from scene motion, enabling better rendering of the scene from novel camera perspectives. Moreover,  For each input frame, we build a 3D point cloud $\mathcal{P} = \{ (\mathbf{x}^{i}, \mathbf{f}^{i})\}$, where $\mathbf{x}^{i}$ represents the 3D location of each 3D point and $\mathbf{f}^{i}$ signifies its appearance feature. Furthermore, we disentangle the process of forecasting future 3D motion to first forecast ego-motion and subsequently predict residual motion for objects. Such disentanglement enables the model to more accurately  forecast 3D motion by focusing on the crucial ego-motion forecasting, where even minor deviation to precisely forecast can lead to significant deviations in visual outcome, then focuses on object motion at various scales.  The following sections detail the process of constructing these point clouds and the methodology for forecasting future 3D motion $\mathbf{u}^{i}$, aiming to displace each 3D point to where it is forecasted to be in the future.

\subsection{Constructing 3D Point Cloud}
\label{sec:fg_decomposition}
\lessgap{}
As shown in Fig.~\ref{fig:3d_scene_decomposition}, we start by estimating the 3D geometry of the input frame $\mathbf{I}$ using the metric depth estimation methods such as ZoeDepth \cite{bhat2023zoedepth}, which can estimate depth in absolute physical units, e.g. meters. Accurate depth map is crucial for 3D scene representation. The resulting depth map is denoted by $\mathbf{D}$.
\vspace{-10pt}
\subsubsection{Semantic segmentation.}
A major challenge in synthesizing photorealistic future frames is handling dis-occlusions caused by dynamic object motion and transitions to novel views. These dis-occlusions appear as ``holes'' in the future frames, where visual information is missing. Many 3D photography methods \cite{shih20203d, li20233d, jampani2021slide, wang20223d} employ depth-based layering \cite{li2021mine} to address dis-occlusions during view transitions. Despite their efficacy, these methods often struggle with dis-occlusions due to the motion of objects within each depth layer. 

To better address dis-occlusions, we leverage semantic segmentation on the input image to identify potential regions for ``holes'' by segmenting both dynamic category objects and likely foreground category objects. Semantic segmentation is performed on $\mathbf{I}$ to generate binary mask $\mathbf{M}$. This mask distinguishes static/background categories (e.g., building, roads, sidewalk, vegetation) from dynamic/foreground categories (e.g., car, bus, pedestrian, pole). For simplicity, we refer to the former category regions as the background and the later as foreground. Note that binary mask $\mathbf{M}$ specifically identifies the background regions. For a detailed categorization, please check supplementary material.
\lessgap{}\lessgap{}
\subsubsection{Image and Depth Inpainting.} To mitigate dis-occlusions, we leverage $\mathbf{M}$ to inpaint the areas where foreground regions are masked out in both  $\mathbf{I}$ and its depth map $\mathbf{D}$ using the surrounding background as context. Importantly, we ensure that the depth values assigned to the inpainted regions are farther in depth than those pixels belonging to the foreground objects. 
This is given by: 
\vspace{-1pt}
\begin{equation}
\mathbf{I}^{\overline{\text{BG}}}, \mathbf{D}^{\overline{\text{BG}}} = \text{Inpaint}(\mathbf{I} \odot \mathbf{M}, \mathbf{D} \odot \mathbf{M}, \mathbf{M}),
\lessgap{}
\lessgap{}
\end{equation}
\vspace{-1pt}
where $\odot$ is element-wise multiplication. Here, $\mathbf{I}^{\overline{\text{BG}}}$ refers to the inpainted image frame, and $\mathbf{D}^{\overline{\text{BG}}}$ is the inpainted depth map.     
\lessgap{}\lessgap{}
\subsubsection{Feature Encoding.} To enhance rendering quality and minimize artifacts in the synthesized future frames, we employ a 2D feature extraction network. This network processes both the original frame, $\mathbf{I}$ and inpainted frame, $\mathbf{I}^{\overline{\text{BG}}}$, generating features $\mathbf{F}$, and $\mathbf{F}^{\overline{\text{BG}}}$, respectively.

Subsequently, we unproject the features $\mathbf{F}$ from the  original 2D frame into 3D space using the corresponding depth map $\mathbf{D}$. To ensure that visual content is present behind the foreground objects in 3D space, we unproject the inpainted regions, using the corresponding features $\mathbf{F}^{\overline{\text{BG}}}$ and depth map $\mathbf{D}^{\overline{\text{BG}}}$ dedicated to these regions. 
This unprojection process facilitates the creation of point clouds, represented as $\mathcal{P}$:
\lessgap{}
\begin{equation}
\mathcal{P} = \text{Unproject}(\mathbf{F}, \mathbf{D}) \cup  \text{Unproject}(\mathbf{F}^{\overline{\text{BG}}}\diamond [1 - \mathbf{M}], \mathbf{D}^{\overline{\text{BG}}} \diamond [1 - \mathbf{M}]),
\label{eq:Unproject}
\lessgap{}
\end{equation} where \(1- \mathbf{M}\) inverts the binary mask and $\diamond$ operator selectively retains pixel-wise features from \(\mathbf{F}\) for which the corresponding value in \(1- \mathbf{M}\) is $1$. For simplicity, we refer to the set $\{\mathbf{F}, \mathbf{D}, \mathbf{M}\}$ as \textit{original feature layer}, denoted by $\mathcal{F}$ and the set $\{\mathbf{F}^{\overline{\text{BG}}}, \mathbf{D}^{\overline{\text{BG}}}, \mathbf{M}\}$ as \textit{inpainted feature layer} $\mathcal{F}^{\overline{\text{BG}}}$.
Applying the above process to both input frames $\mathbf{I}_{(t-1)}$ and $\mathbf{I}_{(t)}$ results in: \begin{equation}
\begin{aligned}
\mathcal{F}_{(t-1)} &= \{\mathbf{F}_{(t-1)}, \mathbf{D}_{(t-1)}, \mathbf{M}_{(t-1)}\}, 
\mathcal{F}_{(t-1)}^{\overline{\text{BG}}} = \{\mathbf{F}_{(t-1)}^{\overline{\text{BG}}}, \mathbf{D}_{(t-1)}^{\overline{\text{BG}}}, \mathbf{M}_{(t-1)}\}.\\
\mathcal{F}_{(t)} &= \{\mathbf{F}_{(t)}, \mathbf{D}_{(t)}, \mathbf{M}_{(t)}\}, 
\mathcal{F}_{(t)}^{\overline{\text{BG}}} = \{\mathbf{F}_{(t)}^{\overline{\text{BG}}},\mathbf{D}_{(t)}^{\overline{\text{BG}}}, \mathbf{M}_{(t)}\}.\\
\end{aligned}
\end{equation}
And the resultant 3D point clouds $\mathcal{P}_{(t-1)}$ and $\mathcal{P}_{(t)}$ can be obtained from feature layers $\{ \mathcal{F}_{(t-1)}, \mathcal{F}_{(t-1)}^{\overline{\text{BG}}} \}$ and $\{ \mathcal{F}_{(t)}, \mathcal{F}_{(t)}^{\overline{\text{BG}}} \}$ using Unproject operation as mentioned in Eq.~\ref{eq:Unproject}. Next, we outline the methodology for generating a future frame using the point clouds $\mathcal{P}_{(t-1)}$ and $\mathcal{P}_{(t)}$, and the forecasted 3D motion for these point clouds. This is followed by a detailed description of forecasting the 3D motion. 

 \begin{figure*}[t!]
    \centering
    \includegraphics[width=1.0\textwidth]{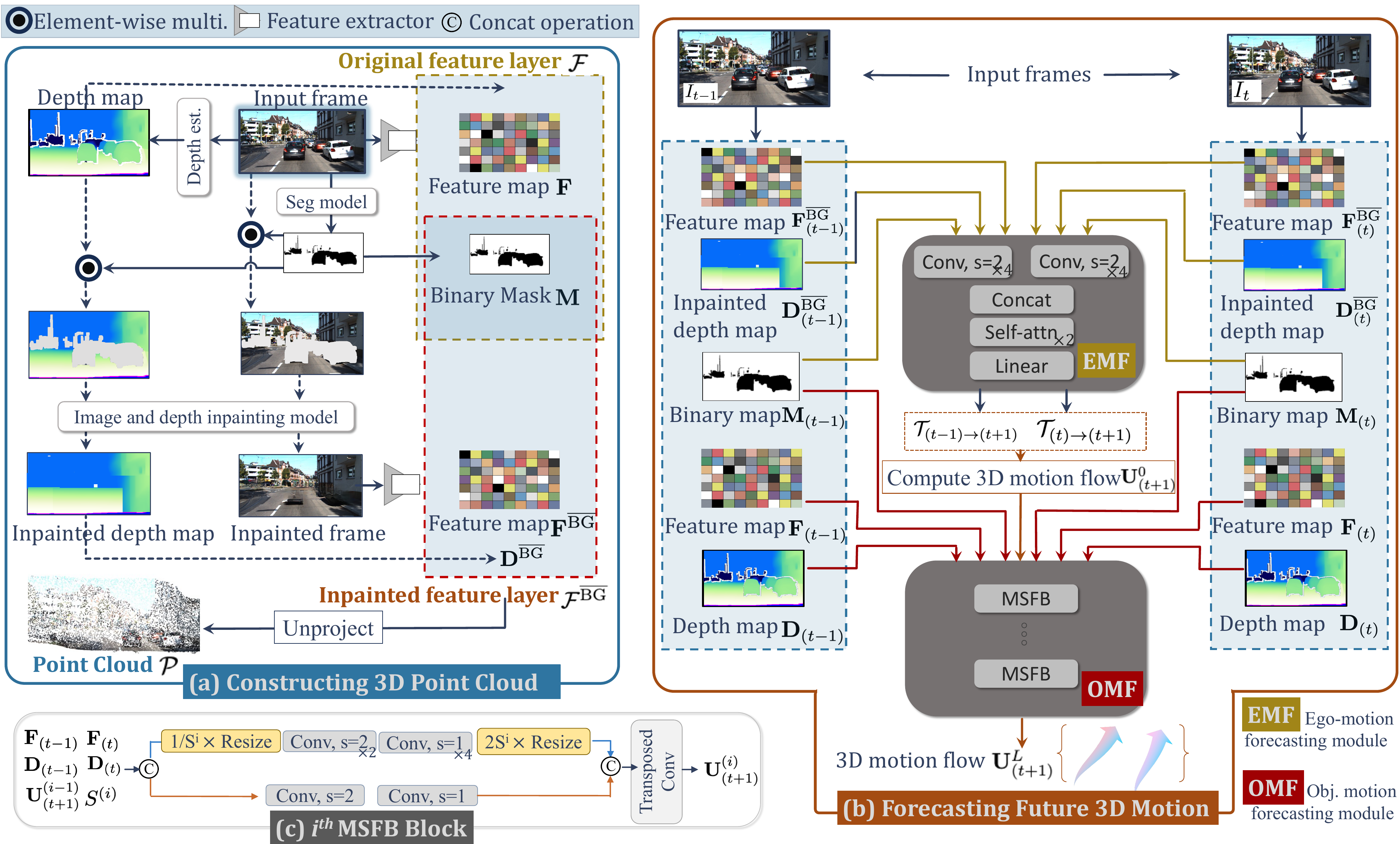}
    \lessgap{}\lessgap{}
    \caption{\textbf{(a) Constructing 3D point cloud.} (1) Estimate the depth map $\mathbf{D}$ from the input image $\mathbf{I}$. (2) Address ``holes’’ in future frames caused by dis-occlusions from dynamic object motion: (i) segment dynamic category (foreground) objects to produce a binary mask $\mathbf{M}$, identifying potential regions for ``holes''. (ii) mask these foreground regions in both input image and depth map, then inpaint them using the background context. (3) Extract features from both original and inpainted frames to produce $\mathbf{F}$ and $\mathbf{F}^{\overline{\text{BG}}}$. (4) Create 3D point cloud $\mathcal{P}$ by unprojecting the 2D features $\mathbf{F}$, $\mathbf{F}^{\overline{\text{BG}}}$ into 3D, using depth maps $\mathbf{D}$, $\mathbf{D}^{\overline{\text{BG}}}$, respectively. For simplicity, we refer to the set $\{\mathbf{F}, \mathbf{D}, \mathbf{M}\}$ as \textit{original feature layer}, denoted by $\mathcal{F}$ and the set of $\{\mathbf{F}^{\overline{\text{BG}}}, \mathbf{D}^{\overline{\text{BG}}}, \mathbf{M}\}$ as \textit{inpainted feature layer} $\mathcal{F}^{\overline{\text{BG}}}$. \textbf{(b) Forecasting future 3D motion.} Given feature layers from past frames, our method forecasts future 3D motion flow in two stages: (1) ego-motion forecasting using the EMF module, which processes the background (static category) across frames using inpainted feature layers $\mathcal{F}_{(t-1)}^{\overline{\text{BG}}}$ and $\mathcal{F}_{(t)}^{\overline{\text{BG}}}$, yielding two relative ego-pose transformations, $\mathcal{T}_{(t-1) \rightarrow (t+1)}$ and $\mathcal{T}_{(t) \rightarrow (t+1)}$. These transformations lead to initial 3D motion flows $\mathbf{u}^{0}_{(t-1) \rightarrow (t+1)}$ and $\mathbf{u}^{0}_{(t) \rightarrow (t+1)}$, referred as $\mathbf{U}^{0}_{(t+1)}$. (2) The OMF module then refines the initial 3D motion flow $\mathbf{U}^{0}_{(t+1)}$ by accounting for foreground object motion, using original and inpainted feature layers to derive the final forecasted 3D motion flow, $\mathbf{U}^{L}_{(t+1)}$, after $L$ MMFB blocks. \textbf{(c) Multi-scale motion flow block (MMFB).} We illustrate the design of a MMFB block here.}
    \label{fig:3d_scene_decomposition}
    \lessgap{}\lessgap{}\lessgap{}
\end{figure*}

\lessgap{}
\subsection{Splatting and Rendering}
\label{sec:rendering_NVS}
\lessgap{}
In order to generate the future frame given a novel camera pose, we leverage the differentiable point-based rendering technique \cite{ravi2020accelerating} that ``splats'' the 3D points in motion adjusted point clouds onto a 2D image plane, synthesizing novel views. 

Given two input frames $\mathbf{I}_{t-1}$ and $\mathbf{I}_{t}$, we have point clouds $\mathcal{P}_{(t-1)}$ and $\mathcal{P}_{(t)}$. Let $\mathbf{u}_{(t-1) \rightarrow (t+1)}$ refer to the forecasted 3D motion flow for 3D points in $\mathcal{P}_{(t-1)}$ and $\mathbf{u}_{(t) \rightarrow (t+1)}$ to the forecasted 3D motion flow for 3D points in $\mathcal{P}_{(t)}$. The method for calculating these motion flows will be explained later.  
The process of generating future frames from a novel viewpoint $\mathbf{K}$ can be summarized in three steps:
\begin{enumerate}
\item Displacement by 3D Motion Flow:
\lessgap{}
    \begin{align}
        \mathcal{P}_{(t-1) \rightarrow (t+1)} = \mathcal{P}_{(t-1)} + \mathbf{u}_{(t-1) \rightarrow (t+1)}, 
        \mathcal{P}_{(t) \rightarrow (t+1)} = \mathcal{P}_{(t)} + \mathbf{u}_{(t) \rightarrow (t+1)},
        \lessgap{}
    \end{align}
\lessgap{}\lessgap{}
\item Scene Rendering via Point-based Splatting:
\lessgap{}
    \begin{align}
        \mathbf{F}_{(t-1) \rightarrow (t+1)} = \text{Render}(\mathcal{P}_{(t-1) \rightarrow (t+1)}, \mathbf{K}), \mathbf{F}_{(t) \rightarrow (t+1)} = \text{Render}(\mathcal{P}_{(t) \rightarrow (t+1)}, \mathbf{K}),
    \end{align}\vspace{-12pt}
    \lessgap{}
\item Future Frame Synthesis:
\lessgap{}
    \begin{equation}
        \hat{\mathbf{I}'}_{(t+1)} = \text{Refine}(\text{Concat}[\mathbf{F}_{(t-1) \rightarrow (t+1)}, \mathbf{F}_{(t) \rightarrow (t+1)}]).
    \end{equation}
\end{enumerate} where ``Concat'' refers to the concatenation of two forecasted features $\mathbf{F}_{(t-1) \rightarrow (t+1)}$, $\mathbf{F}_{(t) \rightarrow (t+1)}$. 
Here, the Render function employs a point-based renderer \cite{ravi2020accelerating}, while the Refine function uses a refinement network based on 2D U-Net architecture \cite{ronneberger2015u}. And $\hat{\mathbf{I}'}_{(t+1)}$ is the forecasted future frame. For long-term forecasts, we can recursively predict future frames $\{ \mathbf{\hat{I'}}_{t+2}, \mathbf{\hat{I'}}_{t+3}, \dots \}$ in a similar manner.
\lessgap{}
\subsection{Forecasting Future 3D Motion}
\label{sec:forecasting}
\lessgap{}
Now, we detail the process of forecasting future 3D motion flows, $\mathbf{u}_{(t-1) \rightarrow (t+1)}$ and $\mathbf{u}_{(t) \rightarrow (t+1)}$, as depicted in Fig.~\ref{fig:3d_scene_decomposition}.   

Most previous video prediction methods focused on forecasting 2D scenes \cite{wu2022optimizing, hu2023dynamic, liu2017video, bei2021learning}, or simplified layered 3D scene. They often rely on pixel-wise backward warping \cite{jaderberg2015spatial} to forecast the next frame from previous frames. 
However, 2D motion forecasting may encounter ambiguities when dealing with occlusions. Additionally, backward warping inherently lacks one-to-one pixel correspondence across frames, resulting in content stretching. In contrast, our 3D motion forecasting mitigates the ambiguities around occluded regions by considering the extra depth information, enabling more accurate estimations of how objects move in and out of each other's paths. To avoid content stretching, we opt for a one-to-one correspondence among 3D points across the future and the past frames.

To accurately forecast future 3D motion, we propose a \textbf{dis-entangled two-stage} approach that first forecasts relative ego-motion and then addresses the residual 3D motion of dynamic objects (e.g., cars, people).  Ego-motion impacts every observed element within the scene, and its interaction with the irregular motion of dynamic objects can lead to inaccuracies in motion prediction.  By disentangling these two components, we can achieve more reliable and detailed forecasts. Moreover, ego-motion forecasting is crucial as it influences all observed elements in the scene. Failure to precisely forecast ego-motion can lead to significant inaccuracies in the predicted scene dynamics, as even minor deviations in camera trajectory prediction can cause large deviations in the visual outcome. 
\lessgap{}\lessgap{}
\subsubsection{Ego-motion forecasting (EMF).} To address this, we design an ego-motion forecasting (EMF) module that infers relative pose changes of the camera from previous frames to future frames.
Our EMF module $E$ takes inpainted features and inpainted depth along with binary mask from inpainted feature layers $\mathcal{F}_{(t-1)}^{\overline{\text{BG}}}$ and $\mathcal{F}_{(t)}^{\overline{\text{BG}}}$ as input, and forecasts the two separate pose transformations $\mathcal{T}_{(t-1) \rightarrow (t+1)}$ and $\mathcal{T}_{(t) \rightarrow (t+1)}$. The transformation $\mathcal{T}$ is parameterized as a rotation matrix $\mathbf{R}$ in the quaternion form ($q_w, q_x, q_y, q_z$) and translation vector $\mathbf{t}$ is ($t_{x}, t_{y}, t_{z}$). The
EMF module can be defined as:
\begin{equation}
\lessgap{}
    \mathcal{T}_{(t-1) \rightarrow (t+1)}, \mathcal{T}_{(t) \rightarrow (t+1)} = E\Big(\mathbf{F}_{(t-1)}^{\overline{\text{BG}}}, \mathbf{D}_{(t-1)}^{\overline{\text{BG}}}, \mathbf{M}_{(t-1)}, \mathbf{F}_{(t)}^{\overline{\text{BG}}}, \mathbf{D}_{(t)}^{\overline{\text{BG}}}, \mathbf{M}_{(t)}\Big).
    \label{eq:egomotion_eqn}
\lessgap{}
\end{equation}
EMF processes past inpainted features $\mathbf{F}^{\overline{\text{BG}}}_{(t-1)}$ and $\mathbf{F}^{\overline{\text{BG}}}_{(t)}$ with corresponding inpainted depths $\mathbf{D}_{(t-1)}^{\overline{\text{BG}}}$ and $\mathbf{D}_{(t)}^{\overline{\text{BG}}}$ through several convolutional layers. After that, these features are concatenated and passed through a series of self-attention blocks \cite{vaswani2017attention}, using masks $\mathbf{M}_{(t-1)}$ and $\mathbf{M}_{(t)}$ to focus on non-inpainted background spatial positions. The self-attention blocks allows the module to focus on relevant features across the temporal sequence, enhancing its ability to predict future ego-motion by identifying patterns and dependencies in the movement of the background across frames. Following the self-attention blocks, the resultant features are input to a linear layer to output the $7$ (4 rotation and 3 translation) parameters for each of the transformations $\mathcal{T}_{(t-1) \rightarrow (t+1)}$ and $\mathcal{T}_{(t) \rightarrow (t+1)}$.
\lessgap{}
\paragraph{Computing 3D motion flow.} Finally, for each of the predicted ego-pose change $\mathcal{T} = (\mathbf{R}, \mathbf{t})$ from Eq.~\ref{eq:egomotion_eqn}, we compute the 3D motion flow $\mathbf{u}$ for every 3D point $p \in \mathcal{P}$ as follows:
\begin{equation}
\lessgap{}
    \mathbf{u} = \mathbf{{R}} \mathbf{x} + \mathbf{t} - \mathbf{x},
\lessgap{}
\end{equation} where $\mathbf{x}$ is the 3D location of point $p$.
\lessgap{}\lessgap{}
\subsubsection{Object motion forecasting (OMF).}
Dynamic objects display a varied nature of motions within a given scene. To handle this diversity, we propose an object motion forecasting module. This module is specifically tailored to predict the residual motion of objects after taking into account the ego-motion, while accounting for the object motions. By doing so, we can achieve a more nuanced and accurate prediction of each object's 3D motion, enhancing the overall realism and coherence of the forecasted 3D scene.

Our object motion forecasting module (OMF) is inspired by the 2D motion forecasting module in DMVFN~\cite{hu2023dynamic}, which preserves spatial resolution and captures significant motion. We thus propose new multi-scale motion flow block (MMFB), constructed using a series of convolutional blocks, with its architecture depicted in Fig.~\ref{fig:3d_scene_decomposition}. Unlike DMVFN, which focuses on appearance features to predict 2D optical flow, our MMFB leverages both appearance features and depth information to estimate future 3D motion. 

Our OMF module consists of $L$ MMFB blocks. As shown in Fig.~\ref{fig:3d_scene_decomposition}, each MMFB block contains two branches for processing the input frame along with the depth map. One branch operates at downsampled scale factor $S$ that focuses on capturing larger-scale motion, while the other branch operates at higher image resolution, enhancing spatial details. Let $\mathbf{u}_{(t) \rightarrow (t+1)}$ encompass the motion field representing the motion of each 3D point. For brevity, let $\mathbf{U}_{(t+1)} = \{\mathbf{u}_{(t-1) \rightarrow (t+1)}$, $\mathbf{u}_{(t) \rightarrow (t+1)}\}$ represent two predicted 3D motion flows from two separate frames. The $i^{th}$ MMFB block, denoted as $F_{MMFB}^{i}$, learns to estimate the target 3D motion flow $\mathbf{U}^{i}_{(t+1)}$ by utilizing the original feature layers $\mathcal{F}_{(t-1)}$ and $\mathcal{F}_{(t)}$, as well as the estimated previous blocks motion flow $\mathbf{U}^{i-1}_{(t+1)}$ and the scale factor $S^{i}$ as input:
\begin{equation}
\lessgap{}
    \mathbf{U}_{(t+1)}^{i} = F_{MMFB}^{i}\Big(\mathbf{F}_{(t-1)}, \mathbf{D}_{(t-1)}, \mathbf{M}_{(t-1)}, \mathbf{F}_{(t)}, \mathbf{D}_{(t)}, \mathbf{M}_{(t)}, \mathbf{U}_{(t+1)}^{i-1}, S^{i}\Big).
\lessgap{}
\end{equation}
 Note that the initial pair of 3D motion flows $\mathbf{U}^{0}_{(t+1)}$ is obtained from  ego-motion forecasting using $\mathcal{T}_{(t-1) \rightarrow (t+1)}, \mathcal{T}_{(t) \rightarrow (t+1)}$, serving as a foundation for subsequent residual motion forecasting of dynamic objects. Finally, after processing through $L$ MMFB layers, OMF obtains the final forecasted 3D motion flow $\mathbf{U}^{L}_{(t+1)} = \{\mathbf{u}_{(t-1) \rightarrow (t+1)}, \mathbf{u}_{(t) \rightarrow (t+1)}\}$.    

\section{Experiments}
\label{sec:experiments}
\lessgap{}
\subsubsection{Dataset and Metric.}
We experiment with two urban datasets. \textbf{KITTI} \cite{geiger2013vision} comprises 28 driving videos with a resolution of $375 \times 1242$, using 24 for training and the remaining 4 for testing. \textbf{Cityscapes} \cite{cordts2016cityscapes} contains 3,475 videos with resolution $1024 \times 2048$. We use 2,945 driving videos for training and 500 videos for testing, all at a resolution of $512 \times 1024$. Our evaluations extend to forecasting up to 5 frames into the future for the KITTI dataset and up to 10 frames for the Cityscapes dataset.  For evaluation, we employ standard error metrics: the Multi-Scale Structural Similarity Index Measure (SSIM) \cite{wang2003multiscale} and the perceptual metric LPIPS \cite{zhang2018unreasonable}, following the conventions of \cite{zhang2022video, le2023waldo}.
\lessgap{}

\subsubsection{Implementation Details.} 
We use the following networks in our framework. 
\begin{itemize}
    \item Pre-trained Networks: we utilize the ZoeDepth \cite{bhat2023zoedepth} method for per-frame depth estimation due to its zero-shot performance across datasets, employing the pre-trained ZoeD-M12-NK model. For segmentation, we incorporate DeepLabV3 \cite{chen2018encoder} pre-trained on the Cityscapes dataset, and the 3D Photo \cite{shih20203d} method for RGB-D inpainting on a per-frame basis.
    \item Trainable Networks: the feature extraction and refinement networks are implemented using the ResNet34 \cite{he2016deep} and 2D U-net architectures \cite{ronneberger2015u}, respectively. The Ego Motion Forecasting (EMF) module leverages a network architecture elaborated in Section~\ref{sec:forecasting}. Furthermore, the 9 ($L$) MMFB blocks within the OMF module, depicted in Figure~\ref{fig:3d_scene_decomposition}, employs a decreasing scaling factor sequence of $[4, 4, 4, 2, 2, 2, 1, 1, 1]$. We provide more details on the architectures in the supplementary material.
\end{itemize}
\lessgap{}
 Training utilizes the Adam optimizer \cite{kingma2014adam} with a cosine annealing strategy and a base learning rate of $10^{-4}$. The training utilizes equal weights of one for both $L1$ and perceptual losses \cite{simonyan2014very}. We train on two A6000 GPUs with a batch size of 4 for 300 epochs. Further, to address issues such as foreground objects becoming semi-transparent due to gaps between samples when the camera zooms in, we use adaptive point-based rendering \cite{ravi2020accelerating} where each 3D point can be rendered by adjusting its point radius proportional to its depth.

\begin{figure*}[t!]
    \centering
    \includegraphics[width=0.98\textwidth]{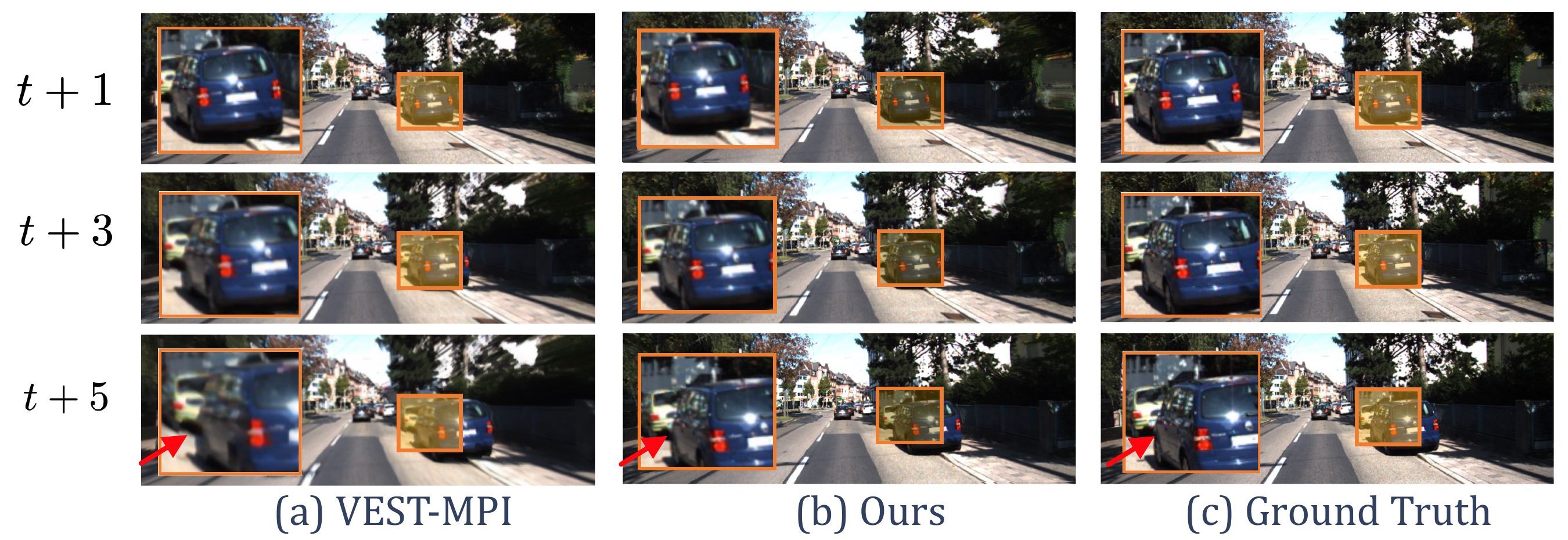}
    \lessgap{}
    \lessgap{}
    \caption{\textbf{Qualitative comparison with VEST-MPI \cite{zhang2022video} on video prediction task (VEST-[S,T]).} The results show that our method produces sharper frames with high-quality motion forecasts, particularly over the long term.}
    \label{fig:VEST-MPI}
    \lessgap{}
    \lessgap{}
\end{figure*}

\begin{table*}[t!]
  \centering
  \resizebox{0.8\textwidth}{!}{ 
  \begin{tabular}{rcc ccc ccc}
    \toprule
    & & & \multicolumn{3}{c}{SSIM ($\times 10^{-2}$)$\uparrow$} & \multicolumn{3}{c}{LPIPS ($\times 10^{-2}$)$\downarrow$}  \\ 
    \cmidrule(lr){4-6} 
    \cmidrule(lr){7-9} 
    Method  & Publication & Inputs & $t+1$ & $t+3$ & $t+5$ & $t+1$ & $t+3$ & $t+5$ \\
    \midrule
     DMVFN \cite{hu2023dynamic} $\rightarrow$ 3D Photo \cite{shih20203d} & CVPR'23 & R+D & 85.4 & 72.9 & 62.6 & 13.0 & 22.7 &  29.2\\
     WALDO \cite{le2023waldo} $\rightarrow$ 3D Photo \cite{shih20203d} & ICCV'23 & R+L+F+D & {85.7} & {75.0} & {68.3} & {12.9} & {19.5} & {24.9}\\

    \midrule
    Ours  & & R+L+D & \bf{86.2} & \bf{76.1} & \bf{69.3} & \bf{12.3} & \bf{17.1} & \bf{21.4}\\
    \bottomrule
  \end{tabular}
  }

  \caption{\textbf{Quantitative results for concurrent video extrapolation in space and time (VEST-[S+T]) on KITTI \cite{geiger2013vision} dataset.}  We use the train-test splits from \cite{le2023waldo} and compare our method with state-of-the-art video prediction techniques, in combination with the 3D Photography method, 3D Photo \cite{shih20203d}, as baselines. `R', `F', `L' and `D' denote the video frames, optical flow, segmentation map, and depth map, respectively. Our method achieves comparable performance with better predictions farther into the future.}
  \label{tab:VEST}
  \largegap{}
\end{table*}



\begin{table}[htbp]
\centering
\begin{minipage}{0.52\textwidth}
    \centering
    \resizebox{1.0\textwidth}{!}{ 
  \begin{tabular}{rc c@{\hspace{0.5em}}c ccc}
    \toprule
    \multicolumn{2}{c}{Extrapolation} & \multicolumn{2}{c}{In space only} & \multicolumn{3}{c}{In time only}\\
    \cmidrule(lr){3-4}
    \cmidrule(lr){5-7}
    & & \multirow{2}{*}{LPIPS$\downarrow$} & \multirow{2}{*}{SSIM$\uparrow$} & \multicolumn{3}{c}{LPIPS ($\times 10^{-2}$)$\downarrow$} \\ 
    \cmidrule(lr){5-7} 
    Method & Publication &  &   & $t+1$ & $t+3$ & $t+5$ \\
    \midrule
     LDI \cite{tulsiani2018layer} & ECCV'18 & N/A & 57.2 & \multicolumn{3}{c}{N/A} \\
     MINE \cite{li2021mine} & ICCV'21 & 10.8 & 82.2 & \multicolumn{3}{c}{N/A}  \\
    Tucker et al. \cite{tucker2020single} & CVPR'20 & N/A & 73.3  & \multicolumn{3}{c}{N/A} \\
    PredRNNV2 \cite{wang2022predrnn} & TPAMI'22 & N/A & N/A & 30.8 & 45.7 & 54.2  \\
    VEST-MPI \cite{zhang2022video} & ECCV'22 & \underline{8.5} & \underline{82.5} & \underline{11.5} & \underline{28.8} & \underline{39.1}\\
    \midrule
    Ours &  & \bf{5.2}  & \bf{94.6} & \bf{8.1} & \bf{18.6} & \bf{20.4}\\
    \bottomrule
  \end{tabular}
  }

  \caption{\textbf{Results for independent VEST (VEST-[S,T]) on KITTI dataset.} We follow \cite{zhang2022video} and use the LDI \cite{tulsiani2018layer} train-test splits. }
  \label{tab:VEST-MPI}
\end{minipage}\hfill
\begin{minipage}{0.44\textwidth}
    \centering
    \resizebox{1.00\textwidth}{!}{ 
  \begin{tabular}{lc ccc ccc}
    \toprule
    &  \multicolumn{3}{c}{SSIM ($\times 10^{-2}$)$\uparrow$} & \multicolumn{3}{c}{LPIPS ($\times 10^{-2}$)$\downarrow$}  \\ 
    \cmidrule(lr){2-4} 
    \cmidrule(lr){5-7} 
    Method  &  $t+1$ & $t+3$ & $t+5$ & $t+1$ & $t+3$ & $t+5$ \\
    \midrule
    A) w/ fixed depth & 83.2 & 73.1 & 66.4 & 15.1 & 20.2 & 25.8 \\
    B) w/o EMF & 85.6 & 75.5 & 68.4 & 12.9 & 18.1 & 22.4\\
    C) w/o OMF & 84.6 & 74.8 & 67.9 & 13.5 & 19.1 & 24.3\\
    \midrule
    Full & \bf{86.2} & \bf{76.1} & \bf{69.3} & \bf{12.3} & \bf{17.1} & \bf{21.4}\\
    \bottomrule
  \end{tabular}
  } 
  \caption{\textbf{Ablation study on VEST-[S+T] on KITTI \cite{geiger2013vision}.}}
  \label{tab:ablation}
\end{minipage}
\largegap{}
\vspace{-5pt}
\end{table}

\lessgap{}
\subsection{Baseline Methods}\lessgap{} We compare our approach with several state-of-the-art methods in video prediction (VP) task, novel view synthesis (NVS) and video extrapolation in space and time (VEST). Additionally, we closely analyze the following baselines for further comparison.

\noindent \textbf{WALDO} \cite{zhang2022video}, a video prediction method that employs layered decomposition to split videos into object-centric layers and a background layer. Then, it generates sparse control points for these layers and forecasts these points into the future, and employs warping and occlusion handling to generate the next frame.

\noindent \textbf{DMVFN} \cite{hu2023dynamic}, a video prediction method that employs dynamic 2D flow estimation via multi-scale voxel flow blocks (MVFB) to capture diverse motion cues between adjacent frames. This estimated flow is utilized for pixel-wise backward warping \cite{jaderberg2015spatial} the past frames to generate next frame. 

\noindent \textbf{VEST-MPI} \cite{zhang2022video} method addresses both Video Prediction (VP) and Novel View Synthesis (NVS) tasks concurrently by introducing a generalized multi-plane image representation. It decomposes images into RGBA planes and parameterizes each plane to effectively model the spatial and temporal dynamics.

\noindent \textbf{Experimental setups.} We compare our approach with state-of-the-art (SOTA) methods using three experimental setups, following the conventions established in prior works \cite{zhang2022video, le2023waldo}. Additionally, we present a novel experimental setup and corresponding baselines.
\lessgap{}
\begin{itemize}
\item Independent video extrapolation in space and time (VEST-[S,T])
\item Concurrent video extrapolation in space and time (VEST-[S+T])
\item Video Prediction (VP)
\end{itemize}
\lessgap{}
\lessgap{}
\subsection{Independent VEST (VEST-[S,T])}\lessgap{}
\noindent \textbf{Setup.} Following VEST-MPI \cite{zhang2022video},  we conduct experiments by independently extrapolating in space (novel view synthesis) and in time (video prediction) using the KITTI dataset \cite{geiger2013vision}, denoted as VEST-[S,T].

\noindent \textbf{Baselines.} We compare our approach with VEST-MPI \cite{zhang2022video} and three leading novel view synthesis methods: LDI \cite{tulsiani2018layer}, MINE \cite{li2021mine}, and Tucker et. al \cite{tucker2020single}. 

\noindent \textbf{Results.} As shown in Table~\ref{tab:VEST-MPI}, our method outperforms the baselines on error metrics. Additionally, comparisons in the video prediction task are provided in  Table~\ref{tab:VP}. In Fig.~\ref{fig:VEST-MPI}, our method is compared with VEST-MPI \cite{zhang2022video} on video prediction task. VEST-MPI struggles with in accurate motion forecasting and rendering details of moving objects, leading to blurred dynamic content (denoted by the red arrow), especially in long-term forecasts. This limitation arises from its highly entangled internal representation. In contrast, our approach synthesizes sharper future frames with high-fidelity motion forecasting by disentangled modeling of 3D scene geometry, camera pose and scene motion.

\begin{table*}[t!]
  \centering
  \resizebox{\textwidth}{!}{ 
  \begin{tabular}{rc c c@{\hspace{0.2em}}c  c@{\hspace{0.2em}}c  c@{\hspace{0.2em}}c   c@{\hspace{0.2em}}c  c@{\hspace{0.2em}}c  c@{\hspace{0.2em}}c}
    \toprule
    & & & \multicolumn{6}{c}{Cityscapes ($512 \times 1024$)} & \multicolumn{6}{c}{KITTI ($256 \times 832$)} \\

    \cmidrule(lr){4-9}
    \cmidrule(lr){10-15}
    
    & & & \multicolumn{2}{c}{$t+1$} & \multicolumn{2}{c}{$t+5$} & \multicolumn{2}{c}{$t+10$} & \multicolumn{2}{c}{$t+1$} & \multicolumn{2}{c}{$t+3$} & \multicolumn{2}{c}{$t+5$}\\ 
    Method & Publication & Inputs & SSIM$\uparrow$ & LPIPS$\downarrow$ & SSIM$\uparrow$ & LPIPS$\downarrow$  & SSIM$\uparrow$ & LPIPS$\downarrow$  & SSIM$\uparrow$ & LPIPS$\downarrow$  & SSIM$\uparrow$ & LPIPS$\downarrow$  & SSIM$\uparrow$ & LPIPS$\downarrow$ \\
    \midrule
    PredNet \cite{lotter2016deep} & NeurIPS'17 & R & 84.0 & 26.0 & 75.2 & 36.0 & 66.3 & 52.2 & 56.3 & 55.3 & 51.4 & 58.6 & 47.5 & 62.9 \\
    MCNet \cite{villegas2017decomposing} & ICLR'17 & R & 89.7 & 18.9 & 70.6 & 37.3 & 59.7 & 45.1 & 53.0 & 24.0 & 63.5 & 31.7 & 55.4 & 37.3 \\
    VFlow \cite{liu2017video} & CVPR'17 & R & 83.9 & 17.4 & 71.1 & 28.8 & 63.4 & 36.6 & 53.9 & 32.4 & 46.9 & 37.4 & 42.6 & 41.5 \\
    OMP \cite{wu2020future} & CVPR'20 & R+L & 89.1 & 8.5 & 75.7 & 16.5 & 67.4 & 23.3 & 79.2 & 18.5 & 67.6 & 24.6 & 60.7 & 30.4  \\
    VPVFI \cite{wu2022optimizing} & CVPR'22 & R & 94.5 & 6.4 & 80.4 & 17.8 & 70.0 & 27.8 & 82.7 & 12.3 & 69.5 & 20.3 & 61.1 & 26.4 \\
    CorrWise \cite{geng2022comparing} & CVPR'22 & R & 92.8 & 8.5 & 83.9 & 15.0 & 75.1 & 21.7 & 82.0 & 17.2 & 73.0 & 22.0 & 66.7 & 25.9 \\
    SADM \cite{bei2021learning} & CVPR'21 & R+L+F & \underline{95.9} & 7.6 & 83.5 & 14.9 & N/A & N/A & 83.1 & 14.4 & 72.4 & 24.6 & 64.7 & 31.2 \\
    DMVFN \cite{hu2023dynamic} & CVPR'23 & R & 95.7 & 5.6 & 83.5 & 14.9 & N/A & N/A & \bf{88.5} & \underline{10.7} & 78.0 & 19.3 & 70.5 & 26.0 \\
    WALDO \cite{le2023waldo} & ICCV'23 & R+L+F & 95.7 & \underline{4.9} & \underline{85.4} & \underline{10.5} & \underline{77.1} & \underline{15.8} & 86.7 & 10.8 & {76.6} & \underline{16.3} & {70.2} & \underline{20.6} \\
    \midrule
     VEST-MPI \cite{zhang2022video} & ECCV'22 & R & N/A & N/A & N/A & N/A & N/A & N/A & N/A & 15.6 & N/A & 34.4 & N/A & 44.7 \\
     \midrule
    Ours &  & R+L+D &  \bf{96.4} & \bf{4.6}  & \bf{86.2} & \bf{9.8} & \bf{78.0} & \bf{14.9}  & \underline{87.7} & \bf{10.1} & \bf{77.6} & \bf{15.4} & \bf{71.3} & \bf{19.8}\\
    
    \bottomrule
  \end{tabular}
}
  \caption{\textbf{Comparisons to state-of-the-art video prediction methods on Cityscapes \cite{cordts2016cityscapes} and KITTI \cite{geiger2013vision} datasets.} We compute multi-scale SSIM ($\times 10^{-2}$) and LPIPS ($\times 10^{-2}$) for the future frame evaluation. `R', `F', `L', `I' and `D' denote the video frames, optical flow, segmentation map, instance map and depth map, respectively. `N/A' means not available.}
  \label{tab:VP}
  \largegap{}
\end{table*}

\lessgap{}

\subsection{Concurrent VEST (VEST-[S+T])}\lessgap{}
\noindent \textbf{Setup.} We proposed to evaluate concurrent video extrapolation in space and time on the KITTI dataset \cite{geiger2013vision}, referred to as VEST-[S+T]. We use train-test splits from \cite{le2023waldo} with an image resolution of $256 \times 832$. 

\noindent \textbf{Baselines.} While VEST-MPI \cite{zhang2022video} provides a promising baseline for comparison, the unavailability of pre-trained models and challenges in reproducibility have limited its application in our study. Instead, we use alternate strong baselines:

\noindent \textbf{WALDO} \cite{le2023waldo} $\rightarrow$ \textbf{3D Photo} \cite{shih20203d}: Combining video prediction and single-image novel view synthesis methods, we first use WALDO \cite{le2023waldo}, to generate a future frame. Then, we employ 3D Photo \cite{shih20203d} to convert the future frame into a layered depth image (LDI), rendering it from a desired viewpoint through a constructed mesh. Additionally, we enhance 3D Photo \cite{wang20223d} with the state-of-the-art monocular depth estimator ZoeDepth \cite{bhat2023zoedepth}.

\noindent \textbf{DMVFN} \cite{hu2023dynamic} $\rightarrow$ \textbf{3D Photo} \cite{shih20203d}: Following the same procedure, we use DMVFN instead of WALDO.

\begin{figure*}[t!]
    \centering
    \includegraphics[width=1.0\textwidth]{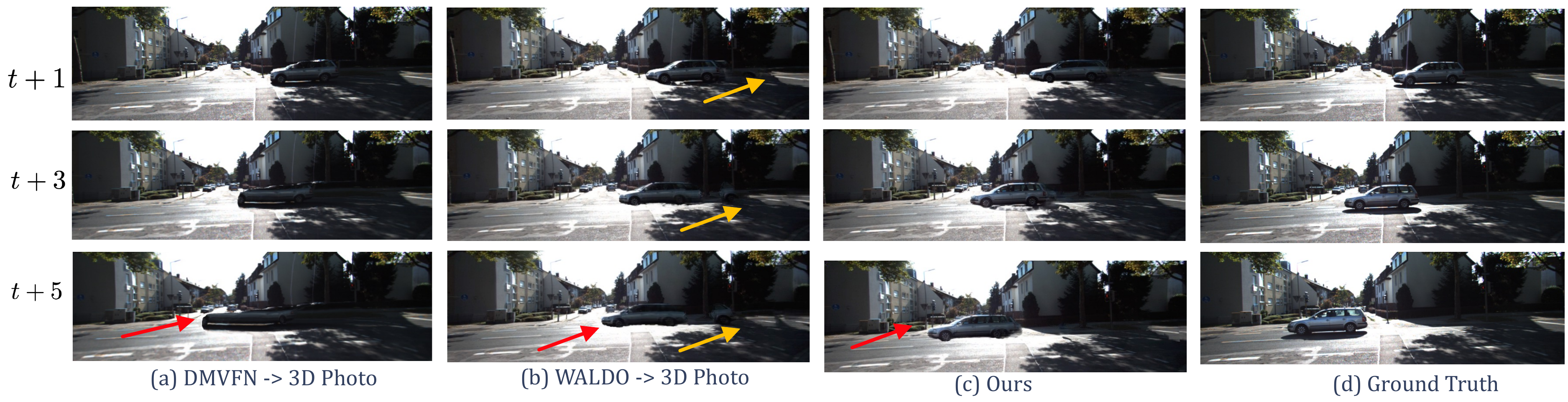}
    \lessgap{}
    \lessgap{}
    \lessgap{}
\lessgap{}
    \caption{\textbf{Qualitative Results for concurrent video extrapolation in space and time (VEST-[S+T])}. (a) The DMVFN \cite{hu2023dynamic} $\rightarrow$ 3D Photo \cite{shih20203d} baseline produces stretching artifacts around the car due to disocclusions caused by the use of 2D flow-based backward warping. (b) In WALDO \cite{le2023waldo} $\rightarrow$ 3D Photo \cite{shih20203d}, \protect\tikz \protect\draw[->, yellow, line width=1pt] (0.0,0.2) -- (0.3,0.3); indicates the inconsistent motion of the road. This occurs due to the layered approach of WALDO \cite{le2023waldo}, where the road and car are mistakenly assigned to the same layer and have similar motion. (c) Our approach mitigates these issues, achieving high-fidelity motion forecasting (indicated by \protect\tikz \protect\draw[->, red, line width=1pt] (0.0,0.2) -- (0.3,0.3);).}
    \label{fig:vest}
    \lessgap{}
    \lessgap{}
\end{figure*}

\noindent \textbf{Results.} Quantitative results in Table~\ref{tab:VEST} show that our approach consistently outperforms prior SOTA methods in error metrics. Notably, our method consistently improves SSIM and LPIPS over long-term forecasts. As depicted in Fig.~\ref{fig:vest}, DMVFN exhibits stretch artifacts due to backward warping and inconsistent occlusion handling resulting from computing flow in the 2D domain. WALDO faces challenges with inaccuracies in layered decomposition, especially for objects moving in different directions but merged into the same layer, creating inconsistent artifacts. Our method effectively addresses these issues through the incorporation of 3D scene geometry and 3D motion modeling using a disentangled approach. 
 
\lessgap{}
\subsection{Video Prediction (VP)} \lessgap{}
In Table~\ref{tab:VP}, we present a quantitative comparison with state-of-the-art (SOTA) video prediction approaches, using train/test splits from \cite{bei2021learning, wang2018video}. Our approach surpasses SOTA methods across all error metrics.


\lessgap{}
\subsection{Ablation Study} \lessgap{} We conduct an ablation study on the KITTI dataset for the VEST-[S+T] task to validate the effectiveness of various proposed system components. Table~\ref{tab:ablation} presents comparisons between our full system and variants: A) Using a fixed depth for both input images, determined by the optimal distance for 3D-2D splatting. Using a fixed depth transforms 3D motion forecasting task into 2D optical flow forecasting task; B) Omitting the ego-motion forecasting (EMF) module and using only the object motion forecasting (OMF) module for forecasting 3D motion; C) using only ego-motion forecasting (EMF) module without object motion forecasting (OMF) module. Without OMF module, synthesized future frame quality degrades as shown in Table~\ref{tab:ablation}. In the absence of EMF module, OMF module can partially compensate for it. However, having an explicit EMF module improves visual quality. 


\section{Discussion and Conclusion}
\lessgap{}
\paragraph{Limitations.} Our approach relies on physical depth estimates and faces challenges with inaccuracy in depth, particularly with thin structures and inconsistencies in depths across frames. These inaccuracies can lead the forecasted results that feature unnatural distortions of thin structures, impacting overall realism of the forecasted frames. Utilizing RGB-D video can be a solution.
\lessgap{}
\lessgap{}
\paragraph{Conclusion.} We present a novel approach for video extrapolation in space and time. By leveraging 3D scene geometry, our method disentangles 3D scene geometry from scene motion. It further disentangles the forecasting of ego-motion from the motion of specific objects, overcoming limitations found in recent approaches. Our approach shows significant improvements over prior state-of-the-art methods in forecasting future videos from novel views.

\section{Acknowledgements}
This material is based upon work supported under the AI Research Institutes program by National Science Foundation and the Institute of Education Sciences, U.S. Department of Education through Award \#2229873 - AI Institute for Transforming Education for Children with Speech and Language Processing Challenges. Any opinions, findings and conclusions or recommendations expressed in this material are those of the author(s) and do not necessarily reflect the views of the National Science Foundation, the Institute of Education Sciences, or the U.S. Department of Education.


%
%
\bibliographystyle{splncs04}
\bibliography{main}
\end{document}


\title{Forecasting Future Videos from Novel Views\\ via Disentangled 3D Scene Representation Supplementary} 

\titlerunning{Abbreviated paper title}



\maketitle
We provide further details of our method as follows:
\begin{enumerate}
    \item Societal impact in Section~\ref{sec:societal_impact}.
    \item Supplementary video details in Section~\ref{sec:suppl_video}.
    \item Detailed categorization of static/background and dynamic/foreground objects in Section~\ref{sec:detailed_category_list}.
    \item Architectural details in Section~\ref{sec:detailed_arch_details}.
\end{enumerate}

\section{Societal Impact}
\label{sec:societal_impact}
This work has the potential to benefit fields such as video prediction and novel view synthesis. The authors believe that this work has small potential negative impacts.

\section{Supplementary Video}
\label{sec:suppl_video}
We have included a supplementary video that presents additional qualitative results alongside baseline comparisons, and it also demonstrates various novel view synthesis and future forecasting capabilities of our method.

\section{Detailed Categorization of Static/Background and Dynamic/Foreground Objects}
\label{sec:detailed_category_list}
In the main manuscript, we discussed leveraging semantic segmentation to distinguish between static/background and dynamic/foreground categories within an image $\mathbf{I}$, resulting in a binary mask $\mathbf{M}$. Here, we provide a detailed breakdown of these categories, elucidating the rationale behind classifying specific objects as either static/background or dynamic/foreground.

\subsection{Static/Background Categories}
The static background encompasses objects and elements within a scene that generally do not change position between consecutive frames or the categories which are not frequently in the foreground. The categories classified under \\static/background include:
\begin{itemize}
    \item \textbf{Building}: Structures including houses, offices, and any stationary constructions.
    \item \textbf{Road}: Surfaces designated for vehicular traffic, including highways and streets.
    \item \textbf{Sidewalk}: Pedestrian paths adjacent to roads.
    \item \textbf{Vegetation}: Trees, grass, and other plant life.
    \item \textbf{Fence}: Barrier structures, including railings.
    \item \textbf{Wall}: Solid structures that limit or enclose areas.
    \item \textbf{Terrain}: Natural land surfaces, including dirt and hills.
    \item \textbf{Sky}: The upper atmosphere viewed from the earth's surface.
\end{itemize}

\subsection{Dynamic/Foreground Categories}
Dynamic/foreground categories represent objects that typically exhibit  change across frames or which are in the foreground more often. These categories include:
\begin{itemize}
    \item \textbf{Person}: Individuals or groups of people.
    \item \textbf{Rider}: Individuals riding bicycles, motorcycles, or other non-enclosed vehicles.
    \item \textbf{Car}: Passenger vehicles including sedans, SUVs, and hatchbacks.
    \item \textbf{Truck}: Larger commercial vehicles used for transporting goods.
    \item \textbf{Bus}: Large passenger vehicles providing public transport services.
    \item \textbf{Train}: Rail vehicles for transporting passengers or cargo.
    \item \textbf{Motorcycle}: Two-wheeled motor vehicles.
    \item \textbf{Bicycle}: Human-powered, pedal-driven vehicles.
    \item \textbf{Pole}: Vertical poles, including lampposts and utility poles.
    \item \textbf{Traffic Light}: Signaling devices situated at road intersections and pedestrian crossings.
    \item \textbf{Traffic Sign}: Signs providing information or instructions to road users.
\end{itemize}

The categorization provided herein aids our semantic segmentation model in effectively identifying potential dis-occlusion regions by distinguishing between the above two categories. 

\begin{table}[b!]
\centering

 \resizebox{1.0\textwidth}{!}{ \begin{tabular}{@{}lllll@{}}
\toprule
Stage & Operation & Input & Output & Output size \\ \midrule
1a' & Conv-stride-2-dil-1-ReLU & $Concat[\mathbf{F}_{(t-1)}^{\overline{\text{BG}}}, \mathbf{D}_{(t-1)}^{\overline{\text{BG}}}, \mathbf{M}_{(t-1)}]$ &$f'_{(t-1)}^{\overline{\text{BG}}}$ & $h/2 \times w/2 \times d'$ \\
1a' & Conv-stride-2-dil-2-ReLU & $f'_{(t-1)}^{\overline{\text{BG}}}$ &$f''_{(t-1)}^{\overline{\text{BG}}}$ & $h/2 \times w/2 \times d'$ \\
1a''' & Conv-stride-2-dil-1-ReLU & $f''_{(t-1)}^{\overline{\text{BG}}}$ &$f'''_{(t-1)}^{\overline{\text{BG}}}$ & $h/4 \times w/4 \times d''$ \\
1a'''' & Conv-stride-2-dil-1-ReLU & $f'''_{(t-1)}^{\overline{\text{BG}}}$ &$f''''_{(t-1)}^{\overline{\text{BG}}}$ & $h/4 \times w/4 \times d''$ \\
1b' & Conv-stride-2-dil-1-ReLU & $Concat[\mathbf{F}_{(t)}^{\overline{\text{BG}}}, \mathbf{D}_{(t)}^{\overline{\text{BG}}}, \mathbf{M}_{(t)}]$ &$f'_{(t-1)}^{\overline{\text{BG}}}$ & $h/2 \times w/2 \times d'$ \\
1b'' & Conv-stride-2-dil-2-ReLU & $f'_{(t)}^{\overline{\text{BG}}}$ &$f''_{(t)}^{\overline{\text{BG}}}$ & $h/2 \times w/2 \times d'$ \\
1b''' & Conv-stride-2-dil-1-ReLU & $f''_{(t)}^{\overline{\text{BG}}}$ &$f'''_{(t)}^{\overline{\text{BG}}}$ & $h/4 \times w/4 \times d''$ \\
1b'''' & Conv-stride-2-dil-1-ReLU & $f'''_{(t)}^{\overline{\text{BG}}}$ &$f''''_{(t)}^{\overline{\text{BG}}}$ & $h/4 \times w/4 \times d''$ \\
2 & 2-Self-Attention Blocks & $Concat[f''''_{(t-1)}^{\overline{\text{BG}}}, f''''_{(t)}^{\overline{\text{BG}}}]$, $\mathbf{M}_{(t-1)}$, $\mathbf{M}_{(t)}$ & $f'''''_{\overline{\text{BG}}}$ & $h/4 \times w/4 \times 2d''$ \\
3 & Linear Layer & $f'''''_{\overline{\text{BG}}}$ & 7 parameters $\times$ 2 (per transformation) \\ \bottomrule
\end{tabular}}
\caption{Detailed Processing Stages of EMF Model}
\label{table:detailed_emf}
\end{table}
\section{Architectural Details of EMF and FMF Module}
\label{sec:detailed_arch_details}
\subsection{EMF Module}
In table \ref{table:detailed_emf} illustrates the processing stages involved in the EMF model for past inpainted features and their corresponding output dimensions. We assume that features $\mathbf{F}_{(t-1)}$ and $\mathbf{F}_{(t)}$ are of size $\mathbb{R}^{h \times w \times d}$, depth maps $\mathbf{D}_{(t-1)}^{\overline{\text{BG}}}$ and $\mathbf{D}_{(t)}^{\overline{\text{BG}}}$, and masks $\mathbf{M}_{(t-1)}$ and $\mathbf{M}_{(t)}$ are of size $\mathbb{R}^{h \times w}$.

\subsection{FMF Module}
The FMF module's key component is the Multi-Scale Feature Block (MSFB), details of which are presented in Figure 3 ``Forecasting future 3D motion''. All convolutional blocks within the MSFB have a stride of 1, and each convolution is followed by a ReLU activation function.


%
%
\bibliographystyle{splncs04}
\bibliography{main}